\documentclass[conference]{IEEEtran}
\IEEEoverridecommandlockouts
\usepackage{cite}
\usepackage{amsmath,amssymb,amsfonts}
\usepackage{algorithmic}
\usepackage{graphicx}
\usepackage{textcomp}
\usepackage{xcolor}
\usepackage{url}
\usepackage{float}
\usepackage{placeins}

\def\BibTeX{{\rm B\kern-.05em{\sc i\kern-.025em b}\kern-.08em
    T\kern-.1667em\lower.7ex\hbox{E}\kern-.125emX}}
\begin{document}

\title{Enhanced Pix2Pix GAN for Visual Defect Removal in UAV-Captured Images}

\author{\IEEEauthorblockN{Volodymyr Rizun}
\IEEEauthorblockA{
\textit{dept. of Artificial Intelligence, IASA}\\ 
\textit{National Technical University of Ukraine
“Igor Sikorsky Kyiv Polytechnic Institute”}\\
Kyiv, Ukraine \\
rizun.volodymyr@lll.kpi.ua}
}

\maketitle

\begin{abstract}   
This paper presents a neural network that effectively removes visual defects from UAV-captured images. It features an enhanced Pix2Pix GAN, specifically engineered to address visual defects in UAV imagery. The method incorporates advanced modifications to the Pix2Pix architecture, targeting prevalent issues such as mode collapse. The suggested method facilitates significant improvements in the quality of defected UAV images, yielding cleaner and more precise visual results. The effectiveness of the proposed approach is demonstrated through evaluation on a custom dataset of aerial photographs, highlighting its capability to refine and restore UAV imagery effectively.
\end{abstract}


\section{Introduction}
A significant challenge faced by contemporary UAV algorithms is their dependence on high-quality input images to generate accurate results. Achieving such image quality is not always feasible, therefore to address this issue, the proposed work introduces a neural network specifically trained to mitigate defects in input images. The network is based on the Pix2Pix architecture but incorporates custom modifications designed to prevent critical errors, thereby ensuring more stable and high-quality outputs.

Let us first take a look at generative adversarial networks (GANs). Since their appearance in 2014 [1], GANs marked a milestone in deep learning and had been a pivotal element in the development of generative AI in particular. Their versatility altogether with the underlying idea made them a powerful framework able to provide state-of-the-art performance, particularly in the domains of image generation and image translation.

GANs also effectively address class imbalance by generating instances for underrepresented classes, improving object detection systems with realistic, degraded images. This work explores this technique and the use of conditional GANs (cGANs) for enhancing visually degraded images.

The original formulation of GANs can be specified as follows. Given a set of objects $X$, namely samples drawn from a data distribution $p_{data}$, the goal is to estimate the probability distribution $p_{data}$. It can be achieved using the following.
\begin{itemize}
    \item The original source of randomness, which is a random variable $\xi$ with a chosen distribution (usually normal or uniform),
    \item The fact that a function of a random variable is also a random variable.
\end{itemize}

In other words, the goal is to construct a function $G$ that takes a random variable with a chosen distribution $p_{\xi}$ and maps it to the given distribution $p_{data}$.

In GANs, the generator, typically a deconvolutional neural network (DeCNN), maps a random variable from a chosen distribution to the data distribution. Optimizing the generator's parameters is challenging due to the limitations of traditional methods like reconstruction loss and likelihood maximization [2]. GANs address these issues by introducing a discriminator, which provides feedback on the data distribution [1]. However, this framework also introduces challenges such as mode collapse, where the generator produces a limited variety of outputs, along with common deep learning issues like vanishing gradients and simplistic loss functions.

Regarding the tasks of image generation and image translation, discriminators are typically implemented as convolutional neural networks (CNNs) [3].

\section{Problem statement}
While GANs provide a general framework for training generative models, they still leave enough space for the exact way for the process of training to be implemented. Originally, the following loss functions were suggested to train the GAN [1].

\begin{equation}\label{eqn:orig_disc_loss}
  \mathcal{L}D = \frac{1}{m} \sum_{i=1}^{m} \left[ \log D(x^{(i)}) + \log(1 - D(G(z^{(i)}))) \right]
\end{equation}
where $\mathcal{L}D$ represents the average logarithmic probability that the discriminator correctly classifies both types of inputs: real samples drawn from $p_{data}$ and generated ones from the generator's distribution $p_{\xi}$.
\begin{equation}\label{eqn:orig_gen_loss}
  \mathcal{L}G = \frac{1}{m} \sum_{i=1}^{m} \left[ \log(1 - D(G(z^{(i)}))) \right]
\end{equation}
where $\mathcal{L}G$ represents the average logarithmic probability that the discriminator correctly identifies the generated samples as fake. Thus, the goal of training a GAN is to adjust parameters of the generator $\theta_{G}$ and the discriminator $\theta_{D}$ to minimize $\mathcal{L}G$ while also maximizing $\mathcal{L}D$. The manner in which these adjustments are made is crucial for ensuring training stability, alongside standard deep learning regularization techniques.

This way, the goal is to construct an optimal schedule for these weight updates to enhance stability and mitigate mode collapse.

\section{Related Work}

This work aims to enhance GAN stability and reduce mode collapse. Key advancements include improvements in loss functions, architectural adjustments, regularization techniques, and data augmentation.

Loss function improvements, such as the Wasserstein GAN (WGAN) by Arjovsky et al. [4], provide smoother gradients and mitigate issues like vanishing gradients. Architectural innovations, including Deep Convolutional GANs (DCGANs) by Radford et al. [5] and Progressive Growing of GANs (ProGAN) by Karras et al. [6], have enhanced stability and image quality. 

Regularization methods like Spectral Normalization [7] and Gradient Penalty [8] further improve stability. Techniques like Differentiable Data Augmentation [9] and Self-Supervised GANs [10] enhance sample diversity and efficiency.

An intriguing approach to addressing mode collapse is the use of hybrid neural networks, which integrate multiple perspectives on the data [11, 12]. Another theoretically promising approach is the use of ensembles, as they offer greater robustness [13]. However, given their typically large size, practical implementation challenges remain significant.

\section{The proposed method}
The discriminator's accuracy should hover around the value of 0.5. However, simply tracking its deviation from this particular value and providing additional iterations to the generator is effectively insufficient. It can lead the generator to adapt merely to the current discriminator's weaknesses, resulting in stagnation. The generator may easily fool the discriminator, which could be trapped in a local optimum.

While selecting an appropriate loss function has proven to be an effective strategy to address this issue by avoiding complex landscapes with numerous saddle points, it does not fully resolve the problem. A robust mechanism to prevent mode collapse by monitoring the progress of both networks is essential. 

This paper introduces such a mechanism by using a relevance threshold, $\nu$, which is a hyperparameter defining the number of most recent epochs. The relevant performance score (RPS) of both networks is calculated in dimensionless units for comparison. Specifically, the idea is to utilize a moving average of performance metrics over the last $\nu$ epochs, which is then normalized by the maximum observed performance during the training. This approach helps in dynamically adjusting the training process by comparing the RPS of the generator and discriminator.

\begin{equation}
    \text{RPS}_{D_{\nu}} = \dfrac{\sum\limits_{i=N - \nu + 1}\limits^{N}\mathcal{L}D_{i}}{\nu \cdot \max \{ \mathcal{L}D_{i} \}}
\end{equation}

\begin{equation}
    \text{RPS}_{G_{\nu}} = \dfrac{\sum\limits_{i=N - \nu + 1}\limits^{N}\mathcal{L}G_{i}}{\nu \cdot \max \{ \mathcal{L}G_{i} \}}
\end{equation}

where $N$ is the total number of epochs so far.

When discrepancies in RPS values exceed a specified threshold, $\varepsilon$, the training process adapts by reallocating more iterations to the underperforming network. 

Formally speaking, if $| \text{RPS}_{G_{\nu}} - \text{RPS}_{D_{\nu}} | > \varepsilon$, then the network with the lower performance score will be allocated additional batches in proportion to the magnitude and direction of the difference \( \text{RPS}_{G_{\nu}} - \text{RPS}_{D_{\nu}} \).

This adjustment helps in maintaining balance and stability between the generator and discriminator, preventing scenarios where one network becomes disproportionately stronger and causes mode collapse.

\begin{figure}[htbp]
    \centerline{\includegraphics{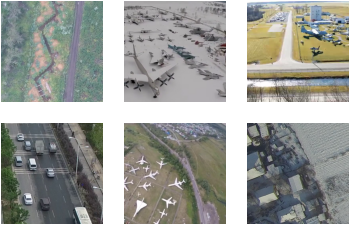}}
    \caption{Sample images (512×512 pixels)}
    \label{fig:examples512}
\end{figure}

\begin{figure}[htbp]
    \centerline{\includegraphics[width=0.7\columnwidth]{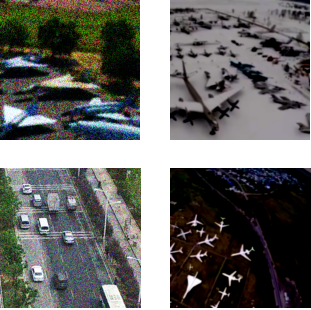}}
    \caption{Degraded images}
    \label{fig:examples256}
\end{figure}

\section{Experiment setup}
In order to test the proposed method, experiments were conducted for some of the most challenging and relevant image translation tasks. For the image translation problem, among cGAN architectures, Pix2Pix [14] was chosen as the baseline model. This model was then upgraded with the proposed adaptive schedule technique for comparison. 

\subsection{Loss functions}
The following expressions were used for the loss functions:

\begin{equation}
    \mathcal{L}_D = -\frac{1}{m} \sum_{i=1}^{m} \sum_{p=1}^{P} \log \left( D_p(x^{(i)}, y^{(i)}) \left(1 - D_p(x^{(i)}, G(x^{(i)})) \right) \right)
\end{equation}
where $m$ is the batch size, $p$ is the number of patches for the PatchGAN discriminator used in Pix2Pix, and $y^{(i)}$ is the corresponding ground truth image for the given input image $x^{(i)}$.

\begin{equation}
    \mathcal{L}G = \mathcal{L}{cGAN}(G, D) + \lambda_{G} \mathcal{L}_{L1}(G)
\end{equation}
where $\mathcal{L}{cGAN}(G, D)$ is the adversarial loss provided to the generator by the discriminator, and $\mathcal{L}_{L1}(G)$ is the reconstruction loss, which is the L1-norm of the pixel-wise difference between the generated image and the corresponding ground truth image. They are defined as follows:
\begin{equation}
    \mathcal{L}{cGAN}(G, D) = -\frac{1}{m} \sum_{i=1}^{m} \sum_{p=1}^{P} \log D_p(x^{(i)}, G(x^{(i)}))
\end{equation}
\begin{equation}
    \mathcal{L}_{L1}(G) = \dfrac{1}{m}  \sum_{i=1}^{m} {\lVert y^{(i)}  -  G(x^{(i)}) \rVert}_1
\end{equation}

\subsection{Evaluation metric}

Fréchet Inception Distance (FID) [15] was used as an evaluation metric to assess the quality of the generated images. FID measures the distance between the distributions of generated images and real images in a feature space extracted by an Inception network. It is calculated as follows:

\begin{equation}
\text{FID} = |\mu_r - \mu_g|^2_2 + \text{Tr}\left(\Sigma_r + \Sigma_g - 2(\Sigma_r \Sigma_g)^{1/2}\right)
\end{equation}

where $\mu_r$ and $\Sigma_r$ are the mean and covariance of the features of the real images, and $\mu_g$ and $\Sigma_g$ are the mean and covariance of the features of the generated images. Lower FID scores indicate better quality and more realistic generated images.

For the experiments, the specific problem of aerial image enhancement was selected, as such images often exhibit such issues as noise, inadequate lighting conditions, and different types of blur [16]. Although GANs can learn to address some of these issues, they may not be able to resolve all problems, primarily due to mode collapse.

Two custom synthetic datasets were created, featuring images from various aerial perspectives (satellites, aircraft, parachute jumps, UAVs) and diverse vehicles (planes, helicopters, cars, vessels). The dataset in [17] comprises UAV images crucial for the experiments. Additional datasets [18-25] contributed to the evaluation.

The two datasets consist of 1,500 images at 256×256 pixels and 2,100 images at 512×512 pixels, cropped from larger originals. Sample images are shown in Figure 1.

The images were then subjected to a custom stochastic degradation pipeline, which randomly introduced noise, inadequate lighting, or various types of blur. Some of these randomly degraded images are depicted in Figure 2.

\begin{figure}[!t]
    \centering
    \includegraphics[width=0.5\textwidth]{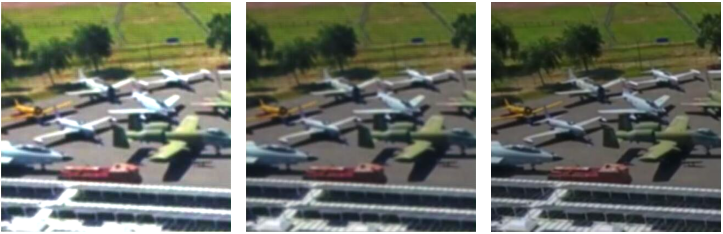}
    \caption{(Left) Input image, (Center) Enhanced by baseline Pix2Pix GAN, (Right) Enhanced by Pix2Pix GAN with the proposed method.}
    \label{fig:results}
\end{figure}

\section{Results}
The models were trained multiple times for 80 epochs on an NVIDIA RTX 3090, with results averaged for consistency. As illustrated in the Figures 4-6, when the Baseline Pix2Pix discriminator outpaces the generator, the latter may become stuck on a plateau while trying to catch up. This allows the discriminator to advance further, leading to a situation where subsequent training becomes nearly impossible. This mode collapse is particularly evident in the FID plot, which shows a spike around the 60th epoch, which is not the case for the proposed modification.

As shown in Figure 3, the results produced by the two GANs highlight distinct differences. The baseline Pix2Pix model managed to address the issue of inadequate lighting but struggled with mitigating blur. In contrast, the proposed method effectively reduced both blur and excessive lighting, thanks to its stable training process which avoids mode collapse.

\begin{figure}[htbp]
    \centerline{\includegraphics[width=0.8\columnwidth]{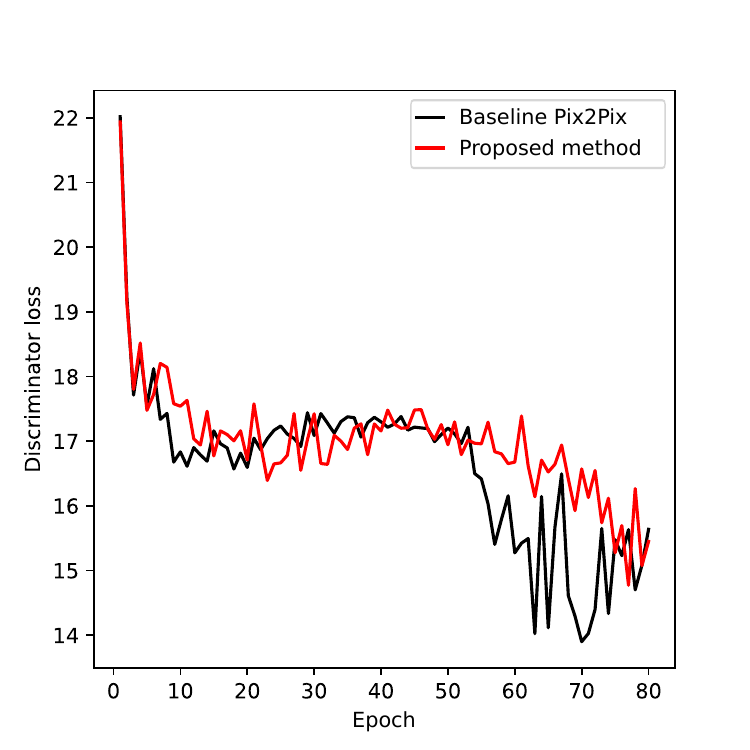}}
    \caption{Discriminator loss dynamics over epochs.}
    \label{fig:disc_loss}
\end{figure}

\begin{figure}[htbp]
    \centerline{\includegraphics[width=0.8\columnwidth]{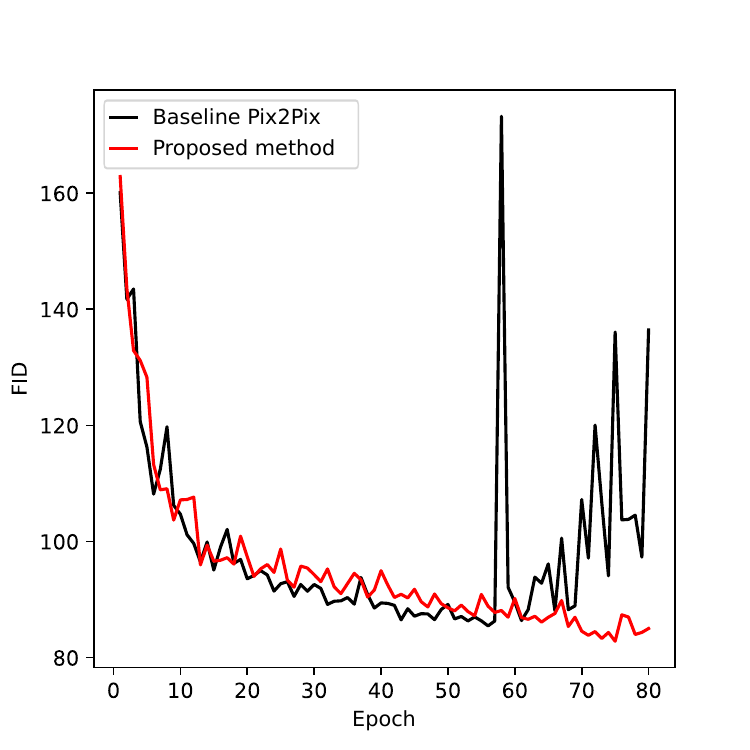}}
    \caption{FID score dynamics over epochs.}
    \label{fig:fid_score}
\end{figure}

\begin{figure}[htbp]
    \centerline{\includegraphics[width=0.8\columnwidth]{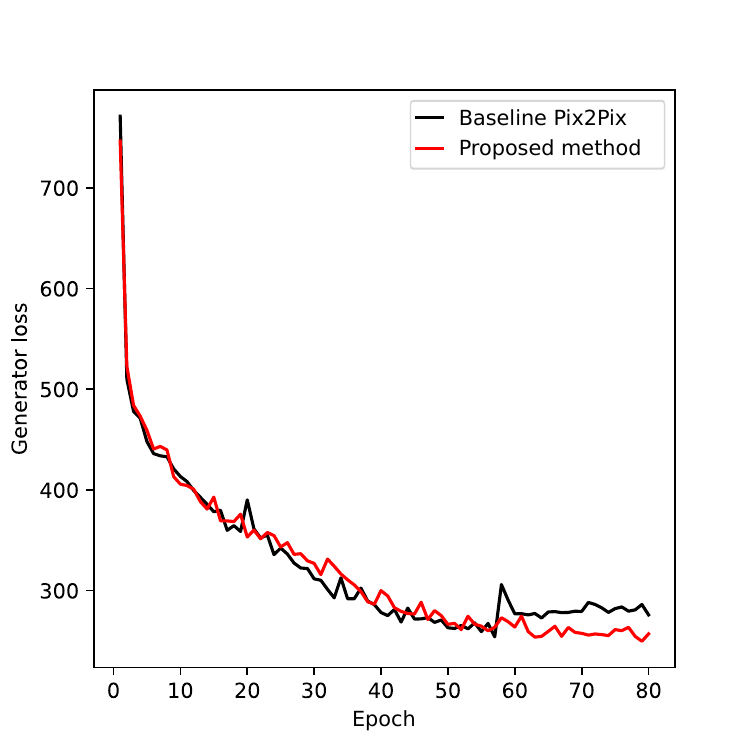}}
    \caption{Generator loss dynamics over epochs.}
    \label{fig:fid_score}
\end{figure}

\section{Conclusion}
Mode collapse and GAN stability are improved by tracking and adapting the update frequency of generator and discriminator weights based on their relative performance. The proposed method ensures balanced progress by continuously monitoring this performance score.

Experiments show that this approach enhances stability and image quality, evidenced by reduced mode collapse and better visual fidelity.


\begin{thebibliography}{00}

\bibitem{b1} I. J. Goodfellow, J. Pouget-Abadie, M. Mirza, B. Xu, D. Warde-Farley, S. Ozair, A. Courville, and Y. Bengio, "Generative Adversarial Networks," 2014. [Online]. Available: \url{https://arxiv.org/abs/1406.2661}

\bibitem{b2} D. J. C. MacKay, "Likelihood-Based Inference for Complex Models," Technical Report, University of Cambridge, 2003.

\bibitem{b3} M. Zgurovsky, V. Sineglazov, and E. Chumachenko, "Classification and Analysis of Topologies of Known Artificial Neurons and Neural Networks," in \textit{Artificial Intelligence Systems Based on Hybrid Neural Networks}, Studies in Computational Intelligence, vol. 904, Springer, Cham, 2021, pp. 1--20, doi: 10.1007/978-3-030-48453-8\_1.


\bibitem{b4} M. Arjovsky, S. Chintala, and L. Bottou, "Wasserstein GAN," 2017. [Online]. Available: \url{https://arxiv.org/abs/1701.07875}

\bibitem{b5} A. Radford, L. Metz, and S. Chintala, "Unsupervised Representation Learning with Deep Convolutional Generative Adversarial Networks," 2016. [Online]. Available: \url{https://arxiv.org/abs/1511.06434}

\bibitem{b6} T. Karras, T. Aila, S. Laine, and J. Lehtinen, "Progressive Growing of GANs for Improved Quality, Stability, and Variation," 2017. [Online]. Available: \url{https://arxiv.org/abs/1710.10196}

\bibitem{b7} T. Miyato, T. Kataoka, M. Koyama, and Y. Yoshida, "Spectral Normalization for Generative Adversarial Networks," in \textit{International Conference on Learning Representations (ICLR)}, 2018.

\bibitem{b8} I. Gulrajani, F. Ahmed, M. Arjovsky, V. Dumoulin, and A. Courville, "Improved Training of Wasserstein GANs," in \textit{Proceedings of the 31st International Conference on Neural Information Processing Systems (NIPS)}, 2017, pp. 5767--5777.

\bibitem{b9} S. Zhao, Z. Liu, J. Lin, J. Zhu, and S. Han, "Differentiable Augmentation for Data-Efficient GAN Training," in \textit{Advances in Neural Information Processing Systems (NeurIPS)}, 2020.

\bibitem{b10} X. Chen, Y. Duan, R. Houthooft, J. Schulman, I. Sutskever, and P. Abbeel, "Infogan: Interpretable Representation Learning by Information Maximizing Generative Adversarial Nets," in \textit{Advances in Neural Information Processing Systems (NIPS)}, 2016, pp. 2172--2180.

\bibitem{b11} V. Sineglazov and A. Kot, "Design of hybrid neural networks of the ensemble structure," \textit{Eastern-European Journal of Enterprise Technologies}, vol. 1, no. 4, pp. 31--45, 2021, doi: 10.15587/1729-4061.2021.225301.

\bibitem{b12} M. Zgurovsky, V. Sineglazov, and E. Chumachenko, "Formation of Hybrid Artificial Neural Networks Topologies," in \textit{Artificial Intelligence Systems Based on Hybrid Neural Networks}, Studies in Computational Intelligence, vol. 904, Springer, Cham, 2021, doi: 10.1007/978-3-030-48453-8\_3.

\bibitem{b13} V. Sineglazov, K. Riazanovskiy, A. Klanovets, E. Chumachenko, and N. Linnik, "Intelligent tuberculosis activity assessment system based on an ensemble of neural networks," \textit{Comput. Biol. Med.}, vol. 147, Aug. 2022, Art. no. 1058003.

\bibitem{b14} P. Isola, J. Zhu, T. Zhou, and A. A. Efros, "Image-to-Image Translation with Conditional Adversarial Networks," in \textit{Proceedings of the IEEE Conference on Computer Vision and Pattern Recognition}, 2017, pp. 1125--1134.

\bibitem{b15} M. Heusel, H. Ramsauer, A. Frome, J. Buchholz, and M. Lucic, "GANs Trained by a Two Time-Scale Update Rule Converge to a Local Nash Equilibrium," in *Proceedings of the 31st International Conference on Neural Information Processing Systems (NeurIPS)*, 2017.

\bibitem{b16} V. M. Sineglazov and S. O. Dolgorukov, "Test bench of UAV navigation equipment," in \textit{2014 IEEE 3rd International Conference on Methods and Systems of Navigation and Motion Control (MSNMC)}, Kyiv, Ukraine, 2014, pp. 108--111, doi: 10.1109/MSNMC.2014.6979743.


\bibitem{b17} Y. Lyu, G. Vosselman, G.-S. Xia, A. Yilmaz, and M. Y. Yang, "UAVid: A semantic segmentation dataset for UAV imagery," \textit{ISPRS Journal of Photogrammetry and Remote Sensing}, vol. 165, pp. 108--119, 2020. [Online]. Available: \url{http://www.sciencedirect.com/science/article/pii/S0924271620301295}. [Accessed: Aug. 9, 2024].

\bibitem{b18} Aerial Image Dataset (AID). [Online]. Available: \url{http://www.escience.cn/people/xinrong/aiddataset.html}

\bibitem{b19} UAV123: A Benchmark Dataset for UAV Object Tracking. [Online]. Available: \url{http://www.cvl.isy.liu.se/research/datasets/uav123/}

\bibitem{b20} SpaceNet: Satellite Imagery Dataset. [Online]. Available: \url{https://spacenet.ai/datasets/}

\bibitem{b21} DeepGlobe Land Cover Classification Dataset. [Online]. Available: \url{https://deepglobe.org/dataset.html}

\bibitem{b22} Airbus Ship Detection Challenge Dataset. [Online]. Available: \url{https://www.kaggle.com/c/airbus-ship-detection/data}

\bibitem{b23} DOTA: Dataset for Object Detection in Aerial Images. [Online]. Available: \url{https://captain-whu.github.io/DOTA/index.html}

\bibitem{b24} xView: A Large-Scale Dataset for Object Detection in Aerial Imagery. [Online]. Available: \url{http://xviewdataset.org/}

\bibitem{b25} UAVDT: A Dataset for UAV Detection and Tracking. [Online]. Available: \url{https://github.com/ucas-vg/UAVDT}

\end{thebibliography}
\end{document}